\documentclass[runningheads]{llncs}
\usepackage{graphicx}

\usepackage{tikz}
\usepackage{comment}
\usepackage{color}

\usepackage[accsupp]{axessibility}  

\usepackage{array}
\usepackage[caption=false,font=normalsize,labelfont=sf,textfont=sf]{subfig}
\usepackage{textcomp}
\usepackage{stfloats}
\usepackage{url}
\usepackage{verbatim}
\usepackage{cite}
\usepackage{mathrsfs}
\usepackage{booktabs}
\usepackage{array}
\usepackage{colortbl}
\usepackage{wrapfig}
\usepackage{multirow}
\usepackage{multicol}
\usepackage[mathscr]{euscript}
\usepackage{ulem}
\usepackage{diagbox}
\usepackage[ruled,linesnumbered]{algorithm2e}

\usepackage[misc]{ifsym} 
\usepackage{bbding}

\makeatletter

\newcommand{\Rmnum}[1]{\expandafter\@slowromancap\romannumeral #1@}
\makeatother

\usepackage{amsmath}
\usepackage{graphicx}
\usepackage{hyperref}
\usepackage{makecell}
\usepackage[latin1]{inputenc}
\usepackage{multirow}
\usepackage{amssymb}

\begin{document}

\pagestyle{headings}
\mainmatter

\title{1st Place Solution for ECCV 2022 OOD-CV Challenge Object Detection Track} 


\titlerunning{ECCV 2022 OOD-CV Challenge Object Detection Track}
\author{Wei Zhao\inst{1,\footnotemark[1]}
\and
Binbin Chen\inst{1,\footnotemark[1]}
\and
Weijie Chen \inst{1,2,\footnotemark[4]}
\and 
Shicai Yang \inst{1}
\and 
Di Xie \inst{1}
\and \\
Shiliang Pu \inst{1}
\and 
Yueting Zhuang \inst{2}
}

\authorrunning{W. Zhao et al.}
%
\institute{Hikvision Research Institute, Hangzhou China
\and
Zhejiang University, Hangzhou, China
\\
\email{\{zhaowei29, chenbinbin8, chenweijie5, yangshicai, xiedi, pushiliang.hri\}@hikvision.com, \{chenweijie, yzhuang\}@zju.edu.cn}
}

\maketitle
\renewcommand{\thefootnote}{\fnsymbol{footnote}}
\footnotetext[1]{Equal contributions}
\footnotetext[4]{Corresponding author}
\footnotetext[5]{https://www.ood-cv.org/}

\begin{abstract}
OOD-CV challenge\footnotemark[5] is an out-of-distribution generalization task. To solve this problem in object detection track, we propose a simple yet effective \textbf{Generalize-then-Adapt (G\&A)} framework, which is composed of a two-stage domain generalization part and a one-stage domain adaptation part. The domain generalization part is implemented by a \textbf{Supervised Model Pretraining} stage using source data for model warm-up and a \textbf{Weakly Semi-Supervised Model Pretraining} stage using both source data with box-level label and auxiliary data (ImageNet-1K) with image-level label for performance boosting. The domain adaptation part is implemented as a \textbf{Source-Free Domain Adaptation} paradigm, which only uses the pre-trained model and the unlabeled target data to further optimize in a self-supervised training manner. The proposed \textbf{G\&A} framework help us achieve the first place on the object detection leaderboard of the OOD-CV challenge. Code will be released in \url{https://github.com/hikvision-research/OOD-CV}.

\end{abstract}

\section{Method} 

\begin{figure}[t!]
  \centering
  \includegraphics[width=\linewidth]{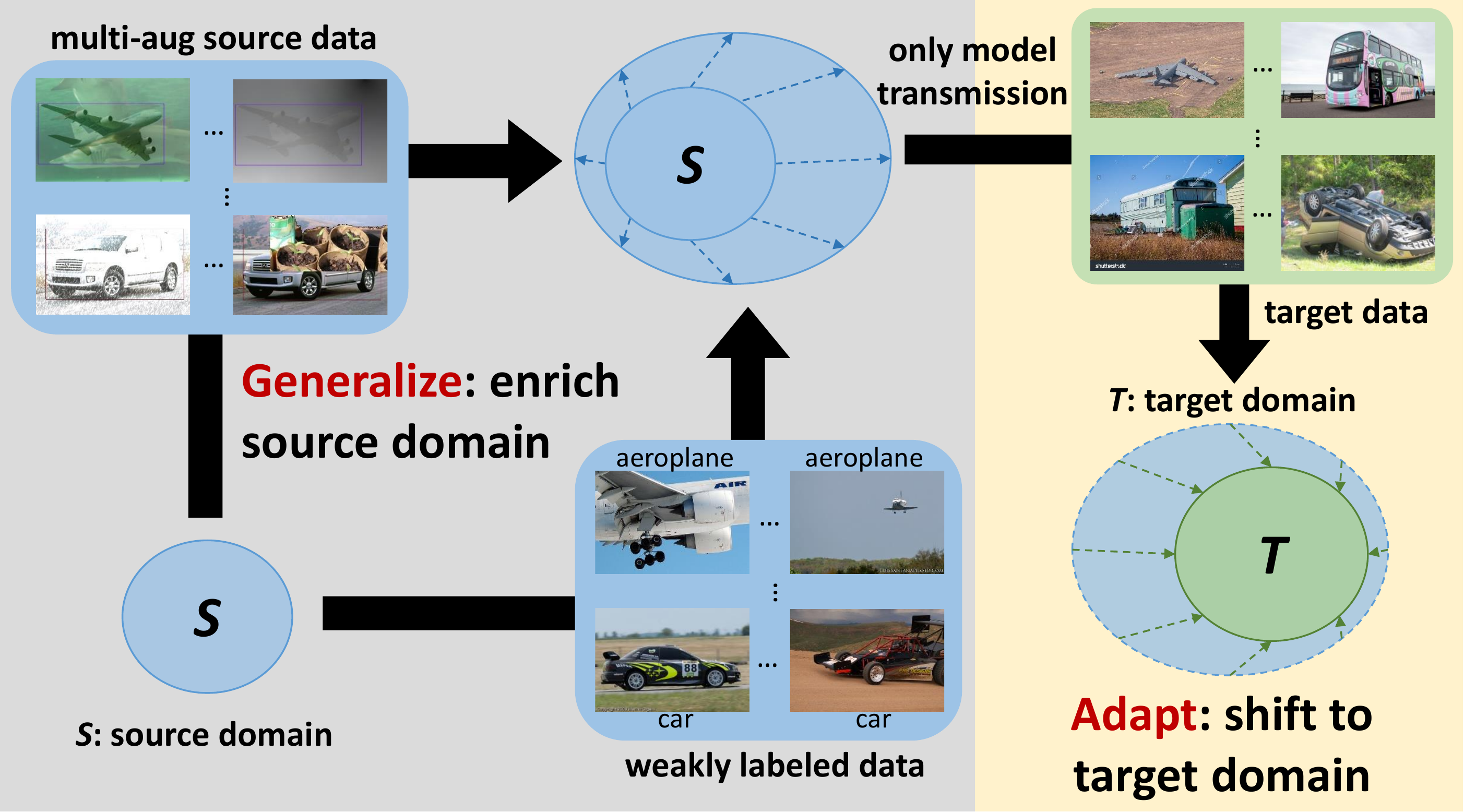}
  \caption{The proposed Generalize-then-Adapt framework, which firstly enriches the source domain as much as possible via strong data augmentation and weakly semi-supervised object detection by exploiting the image-level labeled ImageNet-1K as an auxiliary training data, and then shifts the model performance from the enriched source domains to the given target domain via test-time domain adaptation.}
  \label{GA}
\end{figure}

\begin{figure}[h!]
  \centering
  \includegraphics[width=\linewidth]{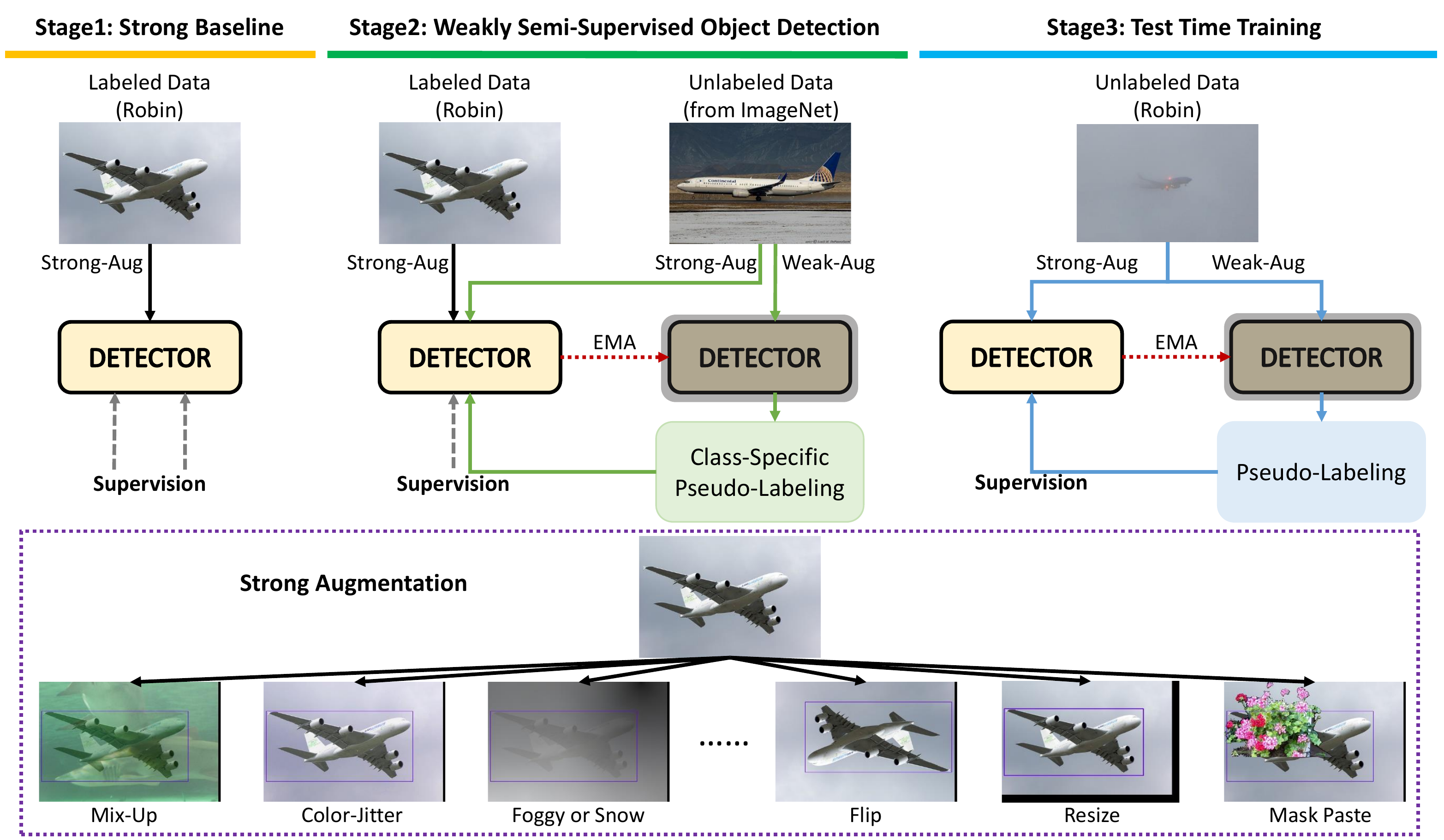}
  \caption{An overview of the proposed Generalize-then-Adapt framework, which is composed of three stages: 1) Strong Baseline (using Robin training set with box-level label), 2) Weakly Semi-Supervised Object Detection (using Robin training set with box-level label and ImageNet-1K with image-level label), 3) Test-Time Training (using unlabeled Robin testing data).}
  \label{PIPELINE}
\end{figure}

To improve the model robustness in the unknown target domains, we propose a simple yet effective \textbf{Generalize-then-Adapt (G\&A)} framework to solve the model degradation problem for object detection under domain shift. Specifically, this framework is composed of a two-stage domain generalization part and a one-stage domain adaptation part: 1) \textbf{Supervised Model Pre-training}. A strong baseline is built against domain shift, which exploits labeled source data with various strong data augmentation strategies so as to simulate potential out-of-distribution data. 2) \textbf{Weakly Semi-Supervised Model Pre-training}. Previous work demonstrates that using extra auxiliary training can further enhance out-of-distribution generalization ability \cite{lin2021semi}. ImageNet-1K \cite{Russakovsky2015ImageNetLS} can be viewed as an auxiliary training data with only image-level label. In this way, the pre-trained object detector in the first stage can be further optimized on the labeled source data (Robin training set \cite{zhao21robin} with box-level label) and the weakly-labeled source data (ImageNet-1K with image-level label), termed Weakly Semi-Supervised Object Detection, which is implemented via a Class-Specific Pseudo-Labeling method in this report. 3) \textbf{Source-Free Domain Adaptation} is utilized for Test-Time Training, which adapts the model to the target domain by merely exploiting the source pre-trained object detector and the unlabeled target data without accessing source data \cite{chen2021self,li2021free}. In this challenge, it is simply implemented as a Mean-Teacher based Self-Training mechanism. The overview of the proposed method is shown in Fig. \ref{PIPELINE}. After integrating Test-Time Augmentation and Model Ensemble strategies, our solution ranks the 1$^{st}$ place on the object detection leaderboard of the OOD-CV challenge \url{https://codalab.lisn.upsaclay.fr/competitions/6784#results}.

\textbf{Our thinking behind this framework}: Compared with conventional domain adaptation methods, which train the source data and the target data jointly (although this paradigm is forbidden in this challenge), the proposed G\&A framework is more feasible in real-world scenarios, which decouples the joint-training paradigm into a domain generalization stage by merely exploiting source data and a test-time domain adaptation stage by merely exploiting target data as shown in Fig.\ref{GA}. G\&A only allows pre-trained model transmission without source data exchange for the sake of avoiding data expansive transmission and data privacy leakage. The generalization step is usually carried out on the server side, while the adaptation step is usually conducted on the client side for model self-evolution. Actually, these two steps can be viewed as an upstream operation and a downstream operation. However, the existing works in OOD community usually focus on the domain generalization step or the test-time domain adaptation step, without taking these two steps into an integration. We hope our solution with superior performance on the leaderboard of this challenge can inspire the community to focus on how to integrate these two steps so as to further resist model degradation under domain shifts.

\section{Implementation Details}

\subsection{Dependencies}
\begin{itemize}
    \item \textbf{Backbone}: \href{https://download.openmmlab.com/mmclassification/v0/convnext/convnext-large_3rdparty_64xb64_in1k_20220124-f8a0ded0.pth}{ConvNext-Large}\cite{liu2022convnet} 
    \item \textbf{Detection Neck}: FPN\cite{lin2017feature}, DyHead\cite{DyHead_CVPR2021}
    \item \textbf{Detection Head}: DDOD\cite{chen2021disentangle}, TOOD\cite{feng2021tood}, VFNet\cite{zhang2020varifocalnet}, Auto-Assign\cite{zhu2020autoassign}, PAA\cite{paa-eccv2020}
    \item \textbf{Ensemble}: WBF\cite{solovyev2021weighted}
    \item \textbf{Dataset}: ROBIN\cite{zhao21robin}, ImageNet-1K\cite{Russakovsky2015ImageNetLS}
    \item \textbf{Weakly-Supervised Semantic Segmentation}: MCTformer\cite{xu2022multi}
\end{itemize}

\subsection{Training Description}

\begin{table}[h!]
\centering
\begin{tabular}{c|c|c}
\hline
AUG & sub-AUG & probability \\
\hline
random resize & / & 1.0 \\
\hline
mask-level copy-paste & / & 0.3 \\
\hline
\multirow{6}{*}{random-choice} & mixup & 0.3 \\
 & blur,noise & 0.3 \\
 & color jitter & 0.3 \\
 & random-erase & 0.3 \\
 & foggy, snow & 0.3 \\
\hline
random horizontal flipping & / & 0.5 \\
\hline
\end{tabular}
\vskip 0.05in
\caption{Strong data augmentation.}
\label{SAUG}
\vskip -0.1in
\end{table}

\subsubsection{Strong baseline.}
A ConvNext-Large based DDOD detection model is trained using the ROBIN training dataset with strong data augmentation. In this stage, the source data is enriched from the original source domain to a wider domain that includes both the original source domain as well as some novel domains to resist domain shifts.

\subsubsection{Weakly semi-supervised object detection.}
To further generalize the object detector, a subset of ImageNet-1K (with the same labels as the 10 categories in ROBIN task) is used as the additional data to enrich the source data with a weakly semi-supervised object detection method based on class-specific pseudo-labeling. Firstly, the pseudo-boxes of ImageNet-1K are generated by forwarding the weak augmented ImageNet-1K dataset to the EMA (Exponential Moving Average) model. Due to the image-level class supervision prior, we only keep the pseudo-boxes belonging to the corresponding image-level labels. Then, the pre-trained model obtained in the first stage is further optimized using the ImageNet-1K dataset and the ROBIN dataset\cite{zhao21robin} with strong data augmentation. The detail about stage2 is shown in Fig. \ref{PIPELINE}-stage2.

\subsubsection{Test-Time Training.}
A source-free domain adaptation based test-time training strategy is used to adapt the model to the testing domain. A simple mean-teacher based self-training mechanism is used in this stage. The weak augmented test data is fed to the EMA model to generate the pseudo label. And then these labels and the corresponding data are used to train the detector with strong data augmentation. The final EMA model is selected for testing.

\subsubsection{Technique details.} 
We build the detector with the backbone of ConvNext-Large and the detection head of DDOD with FPN \cite{lin2017feature}. For data augmentation, we apply mixup\cite{zhang2017mixup}, color jitter, foggy \cite{imagenet-c}, snow \cite{imagenet-c}, resize, horizontal flipping, rotation, blur, noise, random erase\cite{zhong2020random}, and mask-level copy-paste. To obtain more vivid occlusion effects, we introduce MCTformer\cite{xu2022multi} to generate pseudo masks of the salient objects in the subset of ImageNet-1K. Then, these masks are used in the mask-level copy-paste strategy \cite{guo20231st}, which are pasted onto the training images to simulate object occlusion. The data augmentation setting is shown in Table \ref{SAUG}. Note that both mixup and mask-level copy-paste use the ImageNet-1K dataset, and these augmentations are not used in the test-time domain adaptation part. The optimizer for the three training stages described above is AdamW\cite{loshchilov2017decoupled}. The learning rate is 0.0001. The batch size is 2 per GPU. A multi-scale training mechanism is used, and the image height is rescaled between [480, 800] randomly.

\subsection{Testing Description}

In the normal test stage, we only use a single scale (width 1333, height 800). For test-time augmentation (TTA), we use multi-scale and horizontal flipping tricks. The multi-scale trick includes 11 scales, and the image height is evenly ranging from 480 to 800 .

\section{Experimental Results}
\subsection{Ablation Study}

\begin{table}[h!]
\centering
\begin{tabular}{c|ccccc|c|c}
\hline
Method & S1* & S1 & S2 & S3 & TTA & \makecell[c]{OOD-AP50 \\ (Phase-1)} & \makecell[c]{OOD-AP50 \\ (Phase-2)} \\
\hline
\multirow{5}{*}{DDOD} & \checkmark & & & & & 74.43 & / \\
\cline{2-8}
 &  & \checkmark & & & & 83.24 & / \\
\cline{2-8}
 & &\checkmark &\checkmark & & & 85.34 & 68.26 \\
\cline{2-8}
 & &\checkmark &\checkmark & \checkmark & & 85.94 & 71.34 \\
\cline{2-8}
 & &\checkmark &\checkmark & \checkmark & \checkmark & 86.21 & / \\
\hline
\end{tabular}
\vskip 0.05in
\caption{Ablation study. S1, S2 and S3 are short for stage-1, stage-2 and stage-3. S1* only uses weak augmentation in stage-1. TTA is short for test-time augmentation.}
\label{results}
\end{table}

Table \ref{results} shows the ablation study of the proposed method on the phase-1 dataset. The DDOD based ConvNext is the basic object detection framework in this ablation study. The results of S1 and S2 show that the generalization ability of the detection model is gradually enhanced as the training data is enriched. The result of S3 shows that the test-time domain adaptation can effectively shift the model performance from the enriched source domain to the target domain. Besides, the TTA can further improve the model performance.

\subsection{Ensembles And Fusion Strategies}
Experimental evidence shows that the ensemble method can get further performance improvement. We ensemble the predicted boxes from the 15 predictions (8 different models and 7 of them are further performed TTA) by weighted box fusion (WBF). ConvNext-Large\cite{liu2022convnet} is the backbone for all models, and the training settings are exactly the same. Detailed results are shown in Table \ref{ENSEMBLE}. After ensemble, the OOD accuracy (mAP50) is improved from 71.34\% to 73.89\%.

\begin{table}[t!]
\centering
\begin{tabular}{l|cc}
\hline
Model & \makecell[c]{OOD-AP50 \\ (Phase-1)} & \makecell[c]{OOD-AP50 \\ (Phase-2)}\\
\hline
\midrule
DDOD & 85.94 & 71.34\\
+TTA & 86.21 & / \\
\midrule
DDOD+Dyhead1 & 85.42 & / \\
+TTA & 85.92 & / \\
\midrule
DDOD+Dyhead2 & 85.70 & / \\
+TTA & 86.08 & / \\
\midrule
TOOD & 84.93 & / \\
+TTA & 85.92 & / \\
\midrule
Auto-Assign & 83.96 & / \\
+TTA & 84.69 & / \\
\midrule
VFNet & 83.69 & / \\
+TTA & 84.66 & / \\
\midrule
DDOD$\dagger$ & 85.04 & / \\
+TTA & 85.76 & / \\
\midrule
PAA & 85.39 & / \\
\hline
\midrule
Ensemble & 86.67 & 73.89\\
\hline
\end{tabular}
\vskip 0.05in
\caption{Model ensemble performance. $\dagger$ uses random rotation and random vertical flipping except for the strong data augmentations in Table 2. We use WBF\cite{solovyev2021weighted} to fuse the vanilla test results and TTA results from the models of DDOD\cite{chen2021disentangle}, DDOD\cite{chen2021disentangle}+DyHead\cite{DyHead_CVPR2021}, TOOD\cite{feng2021tood}, Auto-Assign\cite{zhu2020autoassign}, VFNet\cite{zhang2020varifocalnet}, DDOD$\dagger$\cite{chen2021disentangle} and PAA\cite{paa-eccv2020}. ``Dyhead1'' and ``Dyhead2'' represent different number of Dyhead blocks.}
\label{ENSEMBLE}
\end{table}

\subsection{Model Complexity Analysis}
\begin{table}[t!]
\centering
\begin{tabular}{c|c}
\hline
Environment & Descriptions \\
\hline
\midrule
system and hardware & Ubuntu20.04+cuda11.3+cudnn8+Tesla V100(32GB)$\times$8 \\
python and pytorch & python=3.8.5+pytorch=1.10 \\
python library & mmcv-full=1.6.1+mmdet=2.25.1+mmcls=0.23.2 \\
\hline
\end{tabular}
\vskip 0.05in
\caption{Language and implementation details.}
\label{IMPLEMENT}
\end{table}

\begin{table}[t!]
\centering
\begin{tabular}{c|c|c}
\hline
 Stage & Time & Descriptions \\
\hline
\midrule
Stage-1 (training) & 13h45min13s & fp16, test-interval=4epoch, total-interval=36epoch \\
Stage-2 (training) & 9h44min26s & fp16, test-interval=1epoch, total-interval=12epoch \\
Stage-3 (training) & 5h36min03s & fp16, test-interval=1epoch, total-interval=6epoch \\
\hline
\midrule
Testing & 26.3 fps & fp32 \\
\hline
\end{tabular}
\vskip 0.05in
\caption{Training/testing time (single model, batch-size=2$\times$8). The test-time augmentation and model ensemble counterparts can be roughly determined from the single-model results, which is omitted here for simplicity. }
\label{TIME}
\end{table}

\begin{table}[t!]
\centering
\begin{tabular}{l|l|l}
\hline
\multirow{3}{*}{ConvNext+DDOD} & Input height & 800 \\
\cline{2-3}
& Flops & 834.61 GFLOPs \\
\cline{2-3}
& Params & 204.71 M \\
\hline
\multirow{3}{*}{ConvNext+DDOD (Multi-Scale)} & Input height & 480:800:32 \\
\cline{2-3}
& Flops & 14687.86 GFLOPs \\
\cline{2-3}
& Params & 204.71 M \\
\hline
\end{tabular}
\vskip 0.05in
\caption{Method complexity and model parameters (single model)}
\label{PARAM}
\end{table}

The environment we use is given in Table \ref{IMPLEMENT}. The training and testing time for DDOD model are shown in Table \ref{TIME}. The proposed solution requires three training stages as shown in Table \ref{TIME}. Therefore, the training time is longer than the base detection framework. However, the test efficiency is the same as the basis detection framework. Please refer to Table \ref{PARAM} for model complexity analysis.

\section{Conclusion}
We thank the organizing committee for providing data \cite{zhao21robin} that allowed us to study the robustness of object detectors under domain shifts.

\bibliographystyle{splncs04}
\bibliography{egbib}
\end{document}